# Minutiae Extraction from Fingerprint Images - a Review

**Roli Bansal[1], Priti Sehgal[2] and Punam Bedi[3]**

**[1] Department of Computer Science,**
**University of Delhi, New Delhi - 110001, India.**

**[2] Reader, Department of Computer Science, Keshav Mahavidyalaya,**
**University of delhi, Pitampura , New Delhi - 110034, India.**

**[3] Associate Professor, Department of Computer Science,**
**University of Delhi, New Delhi - 110001, India.**

## Abstract

Fingerprints are the oldest and most widely used form of biometric identification. Everyone is known to have unique, immutable fingerprints. As most Automatic Fingerprint Recognition Systems are based on local ridge features known as minutiae, marking minutiae accurately and rejecting false ones is very important. However, fingerprint images get degraded and corrupted due to variations in skin and impression conditions. Thus, image enhancement techniques are employed prior to minutiae extraction. A critical step in automatic fingerprint matching is to reliably extract minutiae from the input fingerprint images. This paper presents a review of a large number of techniques present in the literature for extracting fingerprint minutiae. The techniques are broadly classified as those working on binarized images and those that work on gray scale images directly.

## 1. Introduction

Biometrics is the science of uniquely recognizing humans based upon one or more intrinsic physical or behavioral traits. Fingerprints are the most widely used parameter for personal identification amongst all biometrics. Fingerprint identification is commonly employed in forensic science to aid criminal investigations etc. A fingerprint is a unique pattern of ridges and valleys on the surface of a finger of an individual. A ridge is defined as a single curved segment, and a valley is the region between two adjacent ridges. Minutiae points (fig. 1) are the local ridge discontinuities, which are of two types: ridge endings and bifurcations. A good quality image has around 40 to 100 minutiae [1]. It is these minutiae points which are used for determining uniqueness of a fingerprint.

Automated fingerprint recognition and self authentication systems [2] can be categorized as verification or identification systems.

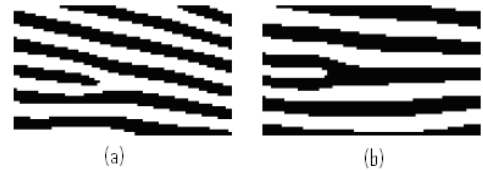

Fig. 1 Minutiae Points.   (a) ridge ending   (b) bifurcation

The verification process either accepts or rejects the user's identity by matching against an existing fingerprint database. In identification, the identity of the user is established using fingerprints. Since accurate matching of fingerprints depends largely on ridge structures, the quality of the fingerprint image is of critical importance. However, in practice, a fingerprint image may not always be well defined due to elements of noise that corrupt the clarity of the ridge structures. This corruption may occur due to variations in skin and impression conditions such as scars, humidity, dirt, and non-uniform contact with the fingerprint capture device. Many algorithms [3-14] have been proposed in the literature for minutia analysis and fingerprint matching and classification for better fingerprint verification and identification. Recently, techniques [15, 16, 17, 18] have been proposed that use other features apart from minutiae for fingerprint recognition. Chen et al [15] propose to reconstruct the fingerprint's orientation field from minutiae and utilize it in the matching stage to improve the system's performance. Cao et al [16] have introduced two novel features to deal with non linear distortion in fingerprints. These features are the finger placement direction and the ridge compatibility. Choi et al [17] proposed to incorporate ridge features like ridge count, ridge length, ridge curvature direction and ridge type together with minutiae to increase the matching performance. Current scientific studies show that application of evolutionary algorithms may improve the performance of biometric systems significantly [19].  There are a number of instances in the literature [20, 21] where evolutionary algorithms are used for matching minutiae of a fingerprint with that of a database of fingerprint images. The results of all such techniques depend on the quality of the input image. Thus, image enhancement techniques are often employed to reduce the noise and to enhance the definition of ridges against valleys so that no spurious minutiae are identified. In fact, matching latent fingerprints from crime scenes is difficult because of their poor quality and the fingerprint matching accuracy is improved by combining manually marked minutiae with automatically extracted ones [22]. Several methods have been proposed for enhancement of fingerprint images which are based on image normalization and Gabor filtering (Hong's algorithm) [1], Directional Fourier filtering [23], Binarization Method [24], enhancement using directional median filter[25], fingerprint image enhancement using filtering techniques[26], image retrieval based on color histogram and textual features[27] and many others[28-32]. The Hong's algorithm inputs a fingerprint image and applies various steps for enhancement. Several other enhancement techniques present in literature are based on fuzzy logic and neural networks [33-40]. Choonwoo et al [41] presented a novel approach to enhance feature extraction for low quality fingerprint images using stochastic resonance (SR). SR refers to a phenomenon where an appropriate amount of noise added to the original signal can increase the signal-to-noise-ratio. Experimental results show that Gaussian noise added to low quality fingerprint images enables the extraction of useful features for biometric identification. The rest of the paper is organized as follows: Section 2 discusses fingerprint features and section 3 explains fingerprint recognition. Section 4 lists the techniques available for minutiae extraction in the literature and finally, section 5 concludes the paper.

## 2. Fingerprint Features

Fingerprint features can be classified into three classes [1]. Level 1 features show macro details of the ridge flow shape, Level 2 features (minutiae point) are discriminative enough for recognition, and Level 3 features (pores) complement the uniqueness of Level 2 features.

### 2.1 Global Ridge Pattern

A fingerprint is a pattern of alternating convex skin called ridges and concave skin called valleys with a spiral-curve-like line shape (fig. 2). There are two types of ridge flows: the pseudo-parallel ridge flows and high-curvature ridge flows which are located around the core point and/or delta point(s). This representation relies on the ridge structure, global landmarks and ridge pattern characteristics. The commonly used global fingerprint features are:

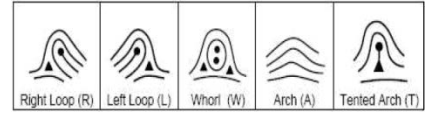

Fig. 2 Global fingerprint ridge patterns

- Singular points – They represent discontinuities in the orientation field. There are two types of singular points as shown in fig. 3. A core is the uppermost of the innermost curving ridge [1], and a delta point is the junction point where three ridge flows meet. They are usually used for fingerprint registration and classification.

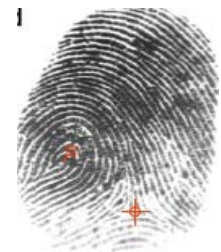

Fig. 3 Singular points (SPs), where "O" and "⊕" denote core and delta, respectively.

- Ridge orientation map – This represents the local direction of the ridge-valley structure. It is commonly utilized for classification, image enhancement and minutia feature verification and filtering.
- Ridge frequency map – It is the reciprocal of the ridge distance in the direction perpendicular to local ridge orientation. It is extensively utilized for contextual filtering of fingerprint images.

### 2.2 Local Ridge Pattern

This is the most widely used and studied fingerprint representation. Local ridge details are the discontinuities of local ridge structure referred to as minutiae. Sir Francis

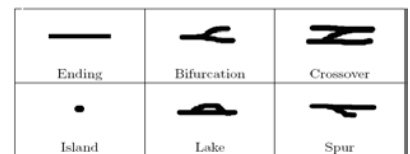

Fig. 4 Some of the common minutiae types

Galton (1822-1922) was the first person who observed the structures and permanence of minutiae. Therefore, minutiae are also called "Galton details". They are used by forensic experts to match two fingerprints. There are about 150 different types of minutiae [3] categorized based on their configuration. Among these minutia types, "ridge ending" and "ridge bifurcation" are the most commonly used, since all the other types of minutiae can be seen as combinations of "ridge endings" and "ridge bifurcations". Some minutiae are shown in fig. 4. The American National Standards Institute-National Institute of Standard and Technology (ANSI-NIST) proposed a minutiae-based fingerprint representation. It includes minutiae location and orientation [42]. Minutia orientation is defined as the direction of the underlying ridge at the minutia location (fig. 5). Minutiae-based fingerprint representation can also assist privacy issues since one cannot reconstruct the original image from using only minutiae information. Actually, minutiae are sufficient to establish fingerprint individuality.

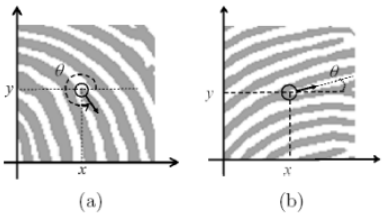

Fig. 5 (a) A ridge ending minutia: (x, y) are the minutia coordinates; θ is the minutia's orientation; (b) A ridge bifurcation minutia: (x, y) are the minutia coordinates; θ is the minutia's orientation.

The minutiae are relatively stable and robust to contrast, image resolutions, and global distortion as compared to other fingerprint representations. However, to extract the minutiae from a poor quality image is not an easy task, although most of the automatic fingerprint recognition systems are designed to use minutiae as the main fingerprint feature for recognition.

## 3 What is Fingerprint Recognition?

The fingerprint recognition [43, 44, 45] problem can be grouped into three sub-domains: fingerprint enrollment, verification and fingerprint identification. In addition, as different from the manual approach for fingerprint recognition by experts, the fingerprint recognition here is referred as AFRS (Automatic Fingerprint Recognition System), which is program-based. Verification is typically used for positive recognition, where the aim is to prevent multiple people from using the same identity. Fingerprint verification is to verify the authenticity of one person by his fingerprint. There is one-to-one comparison in this case. In the identification mode, the system recognizes an individual by searching the templates of all the users in the database for a match. Therefore, the system conducts a one to many comparisons to establish an individual's identity. Both verification and identification use certain techniques for fingerprint matching as indicated in the following subsection.

### 3.1 Techniques for Fingerprint Matching

Various fingerprint matching techniques discussed in literature are as follows:

- Minutiae based technique: Most of the finger-scan technologies are based on Minutiae. Minutia based techniques represent the fingerprint by its local features, like terminations and bifurcations [46-80]. Two fingerprints match if their minutiae points match. This approach has been intensively studied, also is the backbone of the current available fingerprint recognition products.

- Pattern Matching or Ridge Feature Based Techniques: Feature extraction and template generation are based on series of ridges as opposed to discrete points which forms the basis of Pattern Matching Techniques. This includes context aware similarity search techniques applicable to all types of content based image retrieval (CBIR) [81]. The advantage of Pattern Matching techniques [82, 83] over Minutiae based techniques is that minutiae points may be affected by wear and tear and the disadvantages are that these are sensitive to proper placement of finger and need large storage for templates.

- Correlation Based Technique [84, 85, 86] : Let I(Δx, Δy, θ) represent a rotation of the input image I by an angle θ around the origin (usually the image center) and shifted by Δx and Δy pixels in directions x and y, respectively. Then the similarity between the two fingerprint images T and I can be measured as :

$$S(T,I) = \max_{\Delta x, \Delta y, \theta} CC(T, I^{(\Delta x, \Delta y, \theta)}) \qquad (1)$$

where $CC(T, I) = T^{\bar{}}I$ is the cross-correlation between T and I. The cross-correlation is a well known measure of image similarity. It allows us to find the optimal registration. The direct application of eq. (1) rarely leads to acceptable results, mainly due to the following problems:

a) Non-linear distortion makes impressions of the same finger significantly different in terms of global structure; the use of local or block-wise correlation techniques can help to deal with this problem.

b) Skin condition and finger pressure cause image brightness, contrast, and ridge thickness to vary significantly across different impressions. The use of more sophisticated correlation measures may compensate for these problems.

c) The technique is computationally very expensive. Local correlation and correlation in the Fourier domain can improve efficiency.

- Image Based Techniques: Image based techniques try to do matching based on the global features of a whole fingerprint image. It is an advanced and newly emerging method for fingerprint recognition. It is useful to solve some intractable problems of the first approach.

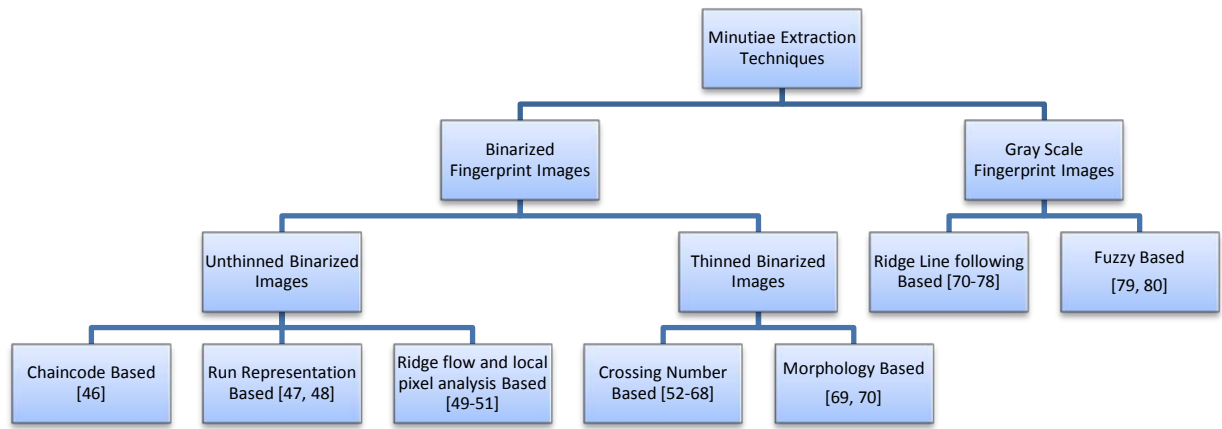

Fig. 6   Classification of Minutiae Extraction Techniques

## 4.      Minutiae Extraction

An accurate representation of the fingerprint image is critical to automatic fingerprint identification systems, because most deployed commercial large-scale systems are dependent on feature-based matching (correlation based techniques have problems as discussed in the previous section). Among all the fingerprint features, minutia point features with corresponding orientation maps are unique enough to discriminate amongst fingerprints robustly; the minutiae feature representation reduces the complex fingerprint recognition problem to a point pattern matching problem. In order to achieve high-accuracy minutiae with varied quality fingerprint images, segmentation algorithm needs to separate foreground from noisy background which includes all ridge-valley regions and not the background. Image enhancement algorithm needs to keep the original ridge flow pattern without altering the singularity, join broken ridges, clean artifacts between pseudo-parallel ridges, and not introduce false information. Finally minutiae detection algorithm needs to locate efficiently and accurately the minutiae points.

There are a lot of minutiae extraction methods available in the literature. We can classify these methods broadly into two categories (fig. 6):

- Those that work on binarized fingerprint images
- Those that work directly on gray-scale fingerprint images.

The following subsections elaborate on the above mentioned categories.

### 4.1      Minutiae detection from binarized fingerprints

A number of binary image based methods are available which detect minutiae by inspecting the localized pixel patterns. They can be further classified into two classes, those that work on unthinned binarized images and those that work on thinned binarized images.

### 4.1.1      Unthinned  Binarized images

Most fingerprint minutia extraction methods are thinning-based where the skeletonization process converts each ridge to one pixel wide. Minutia points are detected by locating the end points and bifurcation points on the thinned ridge skeleton based on the number of neighboring pixels. The end points are selected if they have a single neighbor and the bifurcation points are selected if they have more than two neighbors. However, methods based on thinning are sensitive to noise and the skeleton structure does not conform to intuitive expectation. This category focuses on a binary image based technique of minutiae extraction without a thinning process. The main problem in the minutiae extraction method using thinning processes comes from the fact that minutiae in the skeleton image do not always correspond to true minutiae in the fingerprint image. In fact, a lot of spurious minutiae are observed because of undesired spikes, breaks, and holes.  Therefore, post processing is usually adopted to avoid spurious minutiae, which is based on both statistical and structural information after feature detection. This category discusses three major techniques of minutiae   extraction from unthinned binarized images based on chaincode processing [46], run based methods[47,48] and ridge flow and local pixel analysis based methods [49-51].

#### 4.1.1.1      Chaincode processing

Chaincode representation of object contours is extensively used in document analysis.  Unlike thinned skeletons, the pixel image can be fully recovered from the chaincode of its contour. In this method, the image is scanned from top to bottom and right to left. The transitions from white (background) to black (foreground) are detected. The contour is then traced counterclockwise and expressed as an array of contour elements [46]. Each contour element represents a pixel on the contour. It contains fields for the x, y coordinates of the pixel, the slope or direction of the contour into the pixel, and auxiliary information such as curvature.

In a binary fingerprint image, ridge lines are more than one pixel wide. Tracing a ridge line along its boundary in counterclockwise direction, a termination minutia (ridge ending) is detected when the trace makes a significant left turn. Similarly, a bifurcation minutia (a fork) is detected when the trace makes a significant right turn (fig. 7(a)). Let a vector $P_{in}$ go in to a contour point P and a vector $P_{out}$ go out of P. The computations of $P_{in}$ and $P_{out}$ use several neighboring contour points. This is to avoid local noise and at the same time obtain a better estimation of the vectors using the average of more than one point. The significance of the direction change at P is determined by the angle made between $P_{in}$ and $P_{out}$:

$$\theta = \arccos \frac{P_{in} \cdot P_{out}}{|P_{in}||P_{out}|}$$

(2)

After size normalizations, let the two vectors be $P_{in} = (x1,y1)$ and $P_{out} = (x2,y2)$. Then,

$$\theta = \arccos(x_1 y_1 + x_2 y_2)$$

(3)

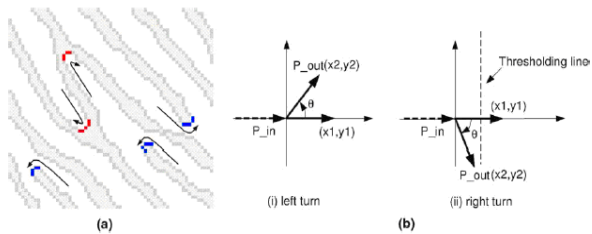

(a)                          (b)

Fig. 7 (a) Minutia location in chaincode contours, the counter wise tracing along the boundary of a ridge line turns left at a termination minutia and turns left at a bifurcation minutia. (b) To calculate the significant turns, the distance between the thresholding line and the y-axis gives a threshold for determining a significant turn [46].

A threshold T is selected so that any significant turn satisfies the condition:

$$x_1 y_1 + x_2 y_2 < T$$

If the vectors are placed in a Cartesian coordinate system with $P_{in}$ along the x-axis (fig. 7(b)), then the threshold T is the x-coordinate of the thresholding line. The turning direction is determined by the sign of sin h since the angle h is always in the range -90 to +90. Therefore,

$x_1 y_2 - x_2 y_1 > 0$ indicates a left turn and

$x_1 y_2 - x_2 y_1 < 0$ indicates a right turn.

This method of direction field generation using chaincode for image enhancement is more efficient and robust for the following reasons: (i) chaincode generation depends on a pre-binarization algorithm, (ii) the adaptive binarization algorithm and the chaincode generation algorithm are both efficient, and (iii) the orientation field is directly computed by tracing the chaincode over a discrete grid. The objective is to attain the ridge orientation for the entire window rather than at every pixel.

### 4.1.1.2    Run Representation

This method results in fast extraction of fingerprint minutiae that are based on the horizontal and vertical run-length encoding from binary images without a computationally expensive thinning process [47, 48]. Fingerprint images are represented by a cascade of runs after run-length encoding. Then runs' adjacency is checked and characteristic runs are detected. But all characteristic runs cannot be true minutiae. So, some geometric constraints are introduced for checking validity of characteristic runs. As shown in fig. 8, the image is preprocessed for enhancement, which is based on the convolution of the image with Gabor filters tuned to the local ridge orientation and frequency. Firstly, the image is segmented [87-90] to extract it from the background. Next, it is normalized so that it has a prespecified mean and variance. After calculating the local orientation and ridge frequency around each pixel, the Gabor filter is applied to each pixel location in the image. As a result the filter enhances the ridges oriented in the direction of local orientation. Hence the filter increases the contrast between the foreground ridges and the background, while effectively reducing noise to set the parameters with respect to the orientation and the frequency, respectively.

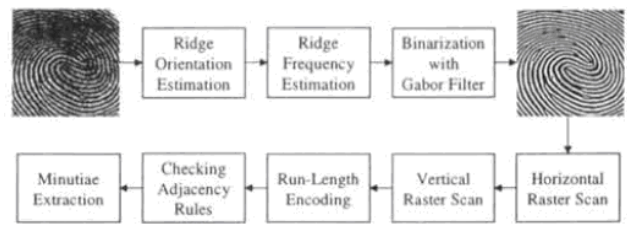

Fig. 8 Block diagram of proposed minutiae extraction algorithm using run-length encoding.

Next, the image is binarized. The simplest way to use image binarization is to choose a threshold value, and classify all pixels with values above this threshold as white, and all other pixels as black. The problem is how to select the correct threshold. In many cases, finding one threshold compatible to the entire image is very difficult, and in many cases even impossible. Therefore, adaptive image binarization is needed where an optimal threshold is chosen for each image area [91, 92].

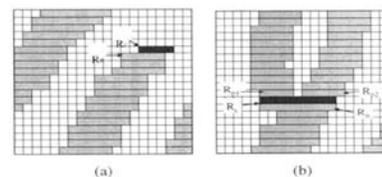

(a)                          (b)

Fig. 9 The minutiae in run representation. (a) Termination in horizontal runs (b) Bifurcation in horizontal runs.

A run-length encoding is an efficient coding scheme for binary or labeled images because it can not only reduce memory space but also speed up image processing time. In the binary image, successive black pixels along the scan line are defined as a run. Generally, a run-length encoding of a binary image is a list of contiguous horizontal runs of black-pixels. For each run a location of the starting pixel of a run and either its length or the location of its ending pixel must be recorded. Fig. 9 shows runs in a binary fingerprint image.

The following five cases are identified:
Case 1 : There are no adjacent runs both on the previous and the next scan line.
Case 2: There are two adjacent runs on the previous and the next scan line.
Case 3: There is one adjacent run on either the previous or the next scan line.
Case 4: There are two adjacent runs on either the previous or the next scan line.
Case 5: There are more than two adjacent runs on either the previous or the next scan line.

The first case means the run is one pixel spot or an isolated line with more than one pixel width. The second case means the run is part of a ridge flow. The third case means the run is a ridge termination, either the starting or the ending points of ridge flow. The fourth case means two runs on the previous scan line are merging or one run is splitting into two runs on the next scan line. Finally, the fifth case has not been considered in the current experiment because a confluence point which is composed of more than two ridge flows is not a minutia in AFRS. The runs in both the third and the fourth cases are called characteristic runs, whereas the runs in the second case are called regular runs. Characteristic runs of the third case correspond to candidates for termination minutiae in a fingerprint image and those of the fourth case stand for bifurcation in a fingerprint image. Some false minutiae are also detected in the process (fig. 10), hence some post processing is necessary for their validation.

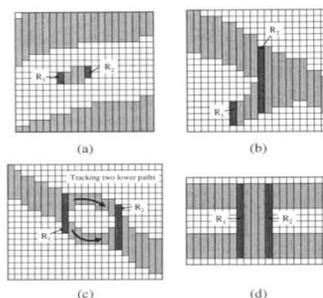

Fig. 10 Examples of false run structures. (a) island (b) spike (c) hole (d) bridge

### 4.1.1.3 Ridge flow and local pixel analysis

Gamassi et al [49] also proposed a square based method to extract minutiae from unthinned binarized images. Around each pixel in the fingerprint image, the method creates a 3x3 square mask and computes the average of pixels. If the average is lesser than 0.25, the pixel is identified as a ridge termination minutiae and if the average is greater than 0.75, the pixel is treated like a bifurcation minutiae. Alibeigi et al [50] further used this method and proposed a hardware scheme based on pipelined architecture for the same. Maddala et al [51] described the implementation and evaluation of an existing fingerprint recognition system developed by the National Institute of Standards and Technology (NIST). The fingerprints are first enhanced and binarized. The binarized image is then scanned both horizontally and vertically using a 2x3 pixel window size to identify ridge endings and bifurcations. A post processing stage is employed to minimize the number of false minutiae.

### 4.1.2 Skeletonization-based Minutiae Extraction (Minutiae Extraction from thinned binarized images with Image Post processing)

Here again the image is preprocessed for enhancement. As explained in the previous section the image is segmented and binarized. Next, the binarized image is thinned. The thinning algorithm removes pixels from ridges until the ridges are one pixel wide [93]. There are other methods also available for thinning [94, 95, 96]. Then the minutiae are extracted from the enhanced, binarized and thinned image. Following the extraction of minutiae, a final image post processing stage is performed to eliminate false minutiae. Most of the techniques in this category are based on the concept of crossing number while some are morphology based.

#### 4.1.2.1 Crossing Number

The most commonly employed method of minutiae extraction in this category is the Crossing Number (CN) concept. A large number of techniques for minutiae extraction available in the literature [52-69] belong to this category.

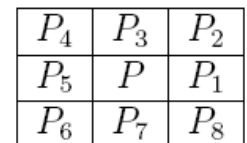

| $P_4$ | $P_3$ | $P_2$ |
|-------|-------|-------|
| $P_5$ | $P$   | $P_1$ |
| $P_6$ | $P_7$ | $P_8$ |

Fig. 11   3X3 neighbourhood

This method is favored over other methods for its computational efficiency and inherent simplicity. This method involves the use of the skeleton image where the ridge flow pattern is eight-connected. The minutiae are extracted by scanning the local neighbourhood of each ridge pixel in the image using a 3X3 window (fig. 11). The CN value is then computed as follows:

$$CN = 0.5 \sum_{i=1}^{8} |P_i - P_{i+1}| \qquad (4)$$

where $P_9 = P_1$. It is defined as half the sum of the differences between pairs of adjacent pixels in the eight-neighbourhood. Using the properties of the CN as shown in fig. 12, the ridge pixel can then be classified as a ridge ending, bifurcation or non-minutiae point. For example, a ridge pixel with a CN of one corresponds to a ridge ending, and a CN of three corresponds to a bifurcation.

| CN | Property |
|----|----------|
| 0 | Isolated point |
| 1 | Ridge ending point |
| 2 | Continuing ridge point |
| 3 | Bifurcation point |
| 4 | Crossing point |

Fig. 12 Properties of crossing number.

Other authors such as Jain et al. [4] have also performed minutiae extraction using the skeleton image. Their approach involves using a 3X3 window to examine the local neighbourhood of each ridge pixel in the image. A pixel is then classified as a ridge ending if it has only one neighbouring ridge pixel in the window, and classified as a bifurcation if it has three neighbouring ridge pixels. Consequently, it can be seen that this approach is very similar to the Crossing Number method.

False minutiae may be introduced into the image due to factors such as noisy images, and image artifacts created by the thinning process. Hence, after the minutiae are extracted, it is necessary to employ a post processing stage in order to validate the minutiae. Fig. 13 illustrates some examples of false minutiae structures, which include the spur, hole, triangle and spike structures [52]. It can be seen that the spur structure generates false ridge endings; whereas both the hole and triangle structures generate false bifurcations. The spike structure creates a false bifurcation and a false ridge ending point.

The majority of the proposed approaches for image post processing in literature [52, 60, 61, and 62] are based on a series of structural rules used to eliminate spurious minutiae. For example, a ridge ending point that is connected to a bifurcation point, and is below a certain threshold distance is eliminated. However, rather than employing a different set of heuristics each time to eliminate a specific type of false minutiae, some approaches incorporate the validation of different types of minutiae into a single algorithm.

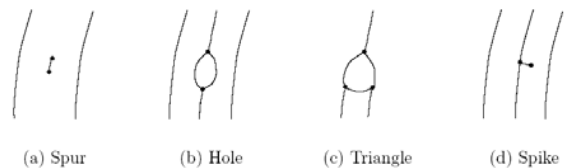

(a) Spur    (b) Hole    (c) Triangle    (d) Spike

Fig. 13 Examples of typical false minutiae structures.

They test the validity of each minutiae point by scanning the skeleton image and examining the local neighbourhood around the minutiae. The algorithm is then able to cancel out false minutiae based on the configuration of the ridge pixels connected to the minutiae point.

In [67], the authors propose fingerprint preprocessing before feature extraction. The preprocessing included obtaining the vertical oriented fingerprint image followed by the core point detection and region of interest selection. Then feature extraction is done in the extracted region of interest image.

Leung et al [68] proposed a neural network based approach to minutiae extraction where preprocessing techniques are first applied to a clean and thinned binary fingerprint ridge structure, which is ready for feature extraction and then a multilayer perceptron network of three layers is trained to extract the minutiae from the thinned fingerprint image.

#### 4.1.2.2 Morphology based

There are minutiae extraction techniques [69, 70] which are based on mathematical morphology. They preprocess the image so as to reduce the effort in the post processing stage. One such technique [70] preprocesses the image with morphological operators to remove spurs, spurious bridges etc. and then uses the morphological Hit or Miss transform to extract true minutiae. Morphological operators are basically shape operators and their composition allows the natural manipulation of shapes for the identification and the composition of objects and object features. The technique develops structuring elements for different types of minutiae present in a fingerprint image to be used by the HMT to extract valid minutiae. Ridge endings are those pixels in an image which have only one neighbour in a 3X3 neighbourhood.

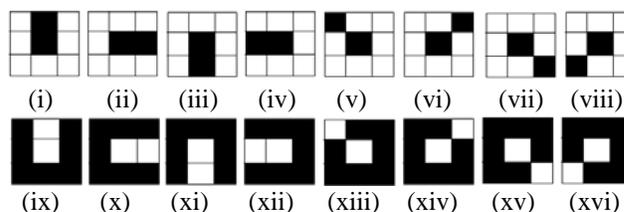

(i)   (ii)   (iii)   (iv)   (v)   (vi)   (vii)   (viii)

(ix)   (x)   (xi)   (xii)   (xiii)   (xiv)   (xv)   (xvi)

Fig. 14 (i) to (viii) The structuring element sequence $J_1 = (J_1{}^1, J_1{}^2, J_1{}^3, J_1{}^4, J_1{}^5, J_1{}^6, J_1{}^7, J_1{}^8)$. (ix) to (xvi) The structuring element sequence $J_2 = (J_2{}^1, J_2{}^2, J_2{}^3, J_2{}^4, J_2{}^5, J_2{}^6, J_2{}^7, J_2{}^8)$.

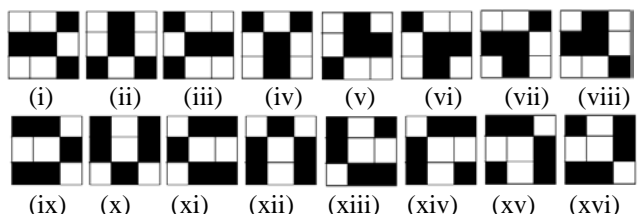

Fig. 15 (i) to (viii) The structuring element sequence $J_1 = (J_1{}^1, J_1{}^2, J_1{}^3, J_1{}^4, J_1{}^5, J_1{}^6, J_1{}^7, J_1{}^8)$. (ix) to (xvi) The structuring element sequence $J_2 = (J_2{}^1, J_2{}^2, J_2{}^3, J_2{}^4, J_2{}^5, J_2{}^6, J_2{}^7, J_2{}^8)$.

The minutiae image M1 containing ridge terminations is given by applying Hit or Miss transform on I by J as follows:

$$M1 \quad = \quad I \quad \otimes \square \tag{5}$$

where, I is the thinned image and J is the sequence of structuring element pairs $(J_1, J_2)$ shown in fig. 14. Ridge bifurcations are those pixels in an image which have only three neighbours in a 3X3 neighbourhood and these neighbours are not adjacent to each other. The minutiae image M2 containing ridge terminations are given by:

$$M2 \quad = \quad I \quad \otimes \square \tag{6}$$

where, I is the thinned image and J is the sequence of structuring element pairs $(J_1, J_2)$ shown in fig. 15. As mentioned earlier, the problem with other techniques is the generation of a large number of spurious minutiae together with true ones whereas this algorithm results in efficient minutiae detection, thereby saving a lot of effort in the post processing stage.

### 4.2 Minutiae Extraction from Gray-Level images

Minutiae detection can also be done directly from gray-level fingerprint images. A number of techniques exist, but it is still a topic of research. Extracting features directly from a gray scale image without binarization and thinning is of great relevance because of the following reasons:

- A lot of information may be lost during binarization process.
- Binarization and thinning are time consuming.
- The aberrations and irregularity of the binary fingerprint image adversely affect the fingerprint thinning procedure and a relatively large number of spurious minutiae are introduced by the binarization thinning operations.
- Most of the binarization techniques prove to be unsatisfactory when applied to low quality images.

### 4.2.1 Minutiae Extraction by following ridge flow lines

Based on the observation that a ridge line is composed of a set of pixels with local maxima along one direction, Maio and Maltoni [71, 72] proposed extracting the minutiae directly from the gray-level image by following the ridge flow lines with the aid of the local orientation field. This method attempts to find a local maximum relative to the cross-section orthogonal to the ridge direction. From any starting point $Pt_s(x_c, y_c)$ with local direction $\theta_c$ in the fingerprint image, a new candidate point $Pt_n(x_n, y_n)$ is obtained by tracing the ridge flow along the $\theta_c$ with fixed step of $\mu$ pixels from $Pt_s(x_c, y_c)$. A new section $\Omega$ containing the point $Pt_n(x_n, y_n)$ is orthogonal to $\theta_c$. The gray-level intensity maxima of $\Omega$ becomes $Pt_s(x_c, y_c)$ to initiate another tracing step. This procedure is iterated till all the minutiae are found. The optimal value for the tracing step $\mu$ and section length $\sigma$ is chosen based on the average width of ridge lines. Jiang et al. [73] improved the method of Maio and Maltoni by choosing dynamically the tracing step $\mu$ according to the change of ridge contrast and bending level. A large step $\mu$ is used when the bending level of the local ridge is low and intensity variations along the ridge direction are small. Otherwise a small step $\mu$ value is used. Instead of tracing a single ridge, Liu et al. [74] proposed tracking a central ridge and the two surrounding valleys simultaneously. In each cross section $\Omega$ a central maximum and two adjacent minima are located at each step, and the ridge following step $\mu$ is dynamically determined based on the distance between the lateral minima from the central maximum. Minutiae are extracted where the relation <minimum, maximum, minimum> is changed. Linear Symmetry (LS) filter in [75, 76] is used to extract the minutiae based on the concept that minutiae are local discontinuities of the LS vector field. Two types of symmetries - parabolic symmetry and linear symmetry are adapted to model and locate the points in the gray-scale image where there is lack of symmetry (fig. 16). A window size of $9 \times 9$ is used to calculate the symmetry filter response. Candidate minutiae points are selected if their responses are above a threshold.

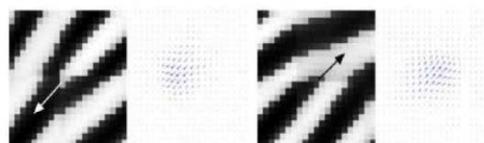

Fig. 16 Symmetry filter response in the minutia point. Left-ridge bifurcation, Right-ridge ending.

Gao et al [77] proposed a minutia extraction method based on Gabor phase. Differing from most existing methods, the approach works in the transform domain of the fingerprint image where, the image is convolved by a Gabor filter, resulting in a complex image. It is then transformed into the amplitude and phase part. A minutiae extractor then extracts minutiae directly from the Gabor phase field. Ratha et al [78] proposed a minutiae extraction algorithm in which the flow direction of ridges is computed by viewing the fingerprint image as a directional textured image. A ridge segmentation algorithm based on a waveform projection is then used to accurately locate the ridges and a thinned ridge image is obtained and smoothed using morphological operators. Finally the minutiae are extracted from the thinned ridges based on the number of crossings and a post processing step applied to remove spurious minutiae.

4.2.2 Fuzzy techniques for minutiae extraction from gray level images

Some fuzzy techniques have also been suggested in literature to extract minutiae from gray scale images directly. Sagar et al [79, 80] proposed that a gray scale image consists of two distinct levels of gray pixels. The darker pixels, constituting the ridges form one such level. The lighter pixels, constituting the valleys and furrows form another such level. Using human linguistics, these two levels of gray can be described as DARK and BRIGHT levels correspondingly. By using fuzzy logic, these two levels are modeled and used along with appropriate fuzzy rules to extract minutiae accurately. For this purpose, rough line thinned structures for both ridges and valleys are obtained. Since bifurcations can be seen as valley endings, the same algorithm that determined ridge endings could be applied to determine valley endings. A 5x5 pixel test window is placed at every point of the line thinned structure. The average value of the 25 pixels is obtained. In addition, the average value of pixels within a 2-pixel border surrounding the test window is also obtained. These two averages form the linguistics variable of brightness. For ridge endings detection, the first average determines the DARK full membership and the second average determines the BRIGHT full membership. This is reversed for the bifurcation detection.

## 5. Conclusions

Image quality is related directly to the ultimate performance of automatic fingerprint authentication systems. Good quality fingerprint images need only minor preprocessing and enhancement for accurate feature detection algorithm. This paper reviewed a large number of techniques described in the literature to extract minutiae from fingerprint images. The approaches are distinguished on the basis of several factors like: the kind of input images they handle i.e. whether binary or gray scale, techniques of binarization and segmentation involved, whether thinning is required or not and the amount of effort required in the post processing stage, if exists. But low quality fingerprint images need preprocessing to increase contrast, and reduce different types of noises as noisy pixels also generate a lot of spurious minutiae as they also get enhanced during the preprocessing steps. Further, more emphasis is to be laid on defining the local criteria, in order to establish the validity of a minutia point, which is particularly useful during fingerprint matching and adopting more sophisticated identification models, for instance extending minutiae definition by including trifurcations, islands, bridges, spurs etc. Also, the paper leads to the further study of the statistical theory of fingerprint minutiae. In particular approaches can be investigated to determine the number of degrees of freedom within a fingerprint population which will give a sound understanding of the statistical uniqueness of fingerprint minutiae.

## References

[1] L. Hong, Y. Wan, and A. K. Jain, "Fingerprint image enhancement: Algorithms and performance evaluation", IEEE Transactions on Pattern Analysis and Machine Intelligence, vol. 20(8), 1988, pp. 777–789.

[2] M. K. Khan, "Fingerprint Biometric-based Self Authentication and Deniable Authentication Schemes for the Electronic World", IETE Technical Review, Volume 26, Issue 3, 2009.

[3] A. K. Jain, L. Hong, and R. Bolle, "On-line fingerprint verification", IEEE Transactions on Pattern Analysis and Machine Intelligence, 19(4), 1997, pp. 302–314.

[4] A. K. Jain, L. Hong, S. Pankanti, and R. Bolle, " An identity authentication system using fingerprints". Proc. IEEE, 85(9), 1997, pp.1365–1388.

[5] A. K. Jain, S. Prabhakar, L. Hong, and S. Pankanti, "Filterbank-based fingerprint matching", Image Processing, IEEE Transactions on, 9(5), 2000, pp. 846–859.

[6] M. Kaur, M. Singh, P.S. Sandhu, "Fingerprint Verification system using Minutiae Verification Technique", Proceedings of world Academy of Science, Engineering and Technology, vol. 36, 2008.

[7] L. H. Thai and N. H. Tam, "Fingerprint recognition using standardized fingerprint model", IJSCI International Journal of Computer Science Issues, vol. 7, issue 3, no. 7, 2010, pp. 11-16.

[8] R. Cappelli, A. Lumini, D. Maio, and D. Maltoni, "Fingerprint classification by directional image partitioning", Pattern Analysis and Machine Intelligence, IEEE Transactions on, 21(5), 2002, pp. 402–421.

[9] R. Cappelli, D. Maio, J. L. Wayman, and A. K. Jain, "Performance evaluation of fingerprint verification systems. IEEE Transactions on Pattern Analysis and Machine Intelligence", 28(1), 2006, pp. 3–18.

[10] S. Pankanti, S. Prabhakar, and A. K. Jain, "On the individuality of fingerprints", IEEE Transactions on Pattern Analysis and Machine Intelligence, 24(8), 2002, pp. 1010–1025.

[11] A. K. Jain, F. Patrick, A. Arun, "Handbook of Biometrics. Springer science and Business media", I edition, 2008 pp. 1-42.

[12] S. Prabhakar, J. Wang, A. K. Jain, S. Pankanti, and R. Bolle, "Minutiae Verification and classification for fingerprint matching". Proc. 15th International Conference Pattern Recognition (ICPR) vol. 1, 2000, pp. 25–29.

[13] B. Bir and T. Xuejun, "Fingerprint indexing based on novel features of minutiae triplets", IEEE Transactions on

Pattern Analysis and Machine Intelligence, 25(5), 2003, pp. 616–622.

[14] Z. Chen and C. H. Kuo, "A topology-based matching algorithm for fingerprint authentication", in proc. IEEE International Carnahan Conference on Security Technology, 1991, pp. 84–87.

[15] F. Chen, J. Zhou and C. Yang, "Reconstructing Orientation Field from Fingerprint Minutiae to Improve Fingerprint Matching Accuracy", IEEE Transactions on Image Processing, vol. 18, no. 7, 2009, pp. 1665-1670.

[16] K. Cao, X. Yang, X. Tao, P. Li, Y. Zang and J. Tian, "Combining features for distorted fingerprint matching", Journal of Network and Computer Applications, vol. 33, 2010, pp. 258-267.

[17] H. Choi, K. Choi and J. Kim, "Fingerprint Matching Incorporating Ridge Features With Minutiae", IEEE Transactions on Information Forensics and Security, vol. 6, no. 2, 2011, pp. 338-345.

[18] S. Kumar D. R., K. B. Raja, R. K. Chhotaray and S. Pattanaik, "DWT Based Fingerprint Recognition using Non Minutiae Features", IJSCI International Journal of Computer Science Issues, vol. 8, issue 2, no. 7, 2011, pp. 237-264.

[19] N. Goranin and A. Cenys, "Evolutionary Algorithms Application Analysis in Biometric systems", Journal of Engineering Science and Technology Review vol. 3, no. 1, 2010, pp. 70-79.

[20] T, V. Le, K. Y. Cheung and M. H. Nguyen, "A Fingerprint Recognizer Using Fuzzy Evolutionary Programming", In Proc. Hawaii International Conference on System Sciences, 2001.

[21] J. Jaam, M. Rebaiaia and A. Hasnah, "A Fingerprint Minutiae Recognition System Based on Genetic Algorithms", The International Arab Journal of Information Technology, vol. 3, no. 3, 2006, pp. 242-248.

[22] A.A. Paulino, A. K. Jain, F. Jianjiang, "Latent Fingerprint Matching: Fusion of Manually Marked and Derived Minutiae," In Proc. 23rd SIBGRAPI Conference on Graphics, Patterns and Images (SIBGRAPI), 2010 , pp.63-70.

[23] B. G. Sherlock, D. M. Monro, and K. Millard, "Fingerprint enhancement by directional Fourier filtering, Vision", IEE Proceedings on Image and Signal Processing, 141(2), 1994, pp. 87–94.

[24] Trier, T. Taxt, "Evaluation of binarisation methods for document images", IEEE Transactions on Pattern Analysis and Machine Intelligence, vol. 17, No. 3, 1995, pp.312–315.

[25] C. Wu, Z. Shi, and V. Govindaraju, "Fingerprint image enhancement method using directional median filter", Biometric Technology for Human Identification, SPIE, volume 5404, 2004, pp. 66–75.

[26] S. Greenberg, M. Aladjem, D. Kogan and I. Dimitrov, "Fingerprint Image Enhancement using Filtering Techniques", Real–Time Imaging vol. 8, 2000, pp. 227–236.

[27] Chuen-Horng Lin, W. Lin, "Image Retrieval System Based on Adaptive Color Histogram and Texture Features", The Computer Journal, 2010.

[28] R. C. Gonzalez and R. E. Woods, "Digital Image Processing", Prentice Hall, Upper Saddle River, NJ, 2002.

[29] D. Maltoni, D. Maio, A. K. Jain, and S. Prabhakar, "Handbook of Fingerprint Recognition", 2003 Springer.

[30] S.K. Oh, J.J. Lee, C.H. Park, B.S. Kim, K.H. Park, "New Fingerprint Image Enhancement Using Directional Filter Bank", Journal of WSCG, vol.11, no.1, 2003.

[31] C. Wu, S. Tulyakov, and V. Govindaraju, "Image quality measures for fingerprint image enhancement", in Proc. International Workshop on Multimedia Content Representation, Classification and Security(MRCS), volume LNCS 4105, 2006, pp. 215–222.

[32] T. Kamei and M. Mizoguchi, "Image filter design for fingerprint enhancement", in Proc. International Symposium on Computer Vision, 1995, pp. 109–114.

[33] M. T. Leung, W. E. Engeler, and P. Frank, "Fingerprint image processing using neural networks", in TENCON 90, IEEE Region 10 Conference on Computer and Communication Systems, vol. 2, 1990, pp. 582–586.

[34] M. Hanmandlu, S. N. Tandon, and A. H. Mir, "A new fuzzy logic based image enhancement", Biomed. Sci. Instrum, vol. 34, 1997, pp. 590–595.

[35] Y.S. Choi and R. Krishnapuram, "A robust approach to image enhancement based on fuzzy logic', IEEE Trans. Image Process., vol 6, no. 6, 1997, pp. 808-825.

[36] M. Natchegael , E. E. Kerre, "Fuzzy techniques in image processing", Springer Verlag, 2000.

[37] A. C. Pais Barreto Marques and A. C. Gay Thome, "A neural network fingerprint segmentation method", in Proc. Fifth International Conference on Hybrid Intelligent Systems, 2006.

[38] M. T. Yildrim, A. Basturk, "A Detail Preerving type-2 Fuzzy Logic Filter for Impulse Noise Removal from Digital Images", Fuzzy Systems Conference, FUZZ-IEEE, 2007.

[39] R. Bansal, P. Sehgal, P. Bedi, "A novel framework for enhancing images corrupted by impulse noise using type-II fuzzy sets", in Proc. IEEE International Conference on Fuzzy Systems and Knowledge Discovery(FSKD'2008) vol. 3, 2008, pp. 266-271.

[40] R. Bansal, Malvika Gaur, Payal Arora, P. Sehgal, P. Bedi, "Fingerprint Image Enhancement Using Type-2 fuzzy sets", in Proc. IEEE International Conference on Fuzzy Systems and Knowledge Discovery(FSKD'2009), Tianjin, China , vol. 3, 2009, pp, 412-417.

[41] C. Ryu, S. G. Kong and H. Kim, "Enhancement of feature extraction for low-quality fingerprint images using stochastic resonance", Pattern Recognition Letters, vol. 32, 2011, pp. 107-113.

[42] A. Bazen and S. Gerez, "Segmentation of fingerprint images", in Proc. Workshop on Circuits Systems and Signal Processing (ProRISC 2001), pp. 276–280.

[43] A. J. Willis and L. Myers, "A cost-effective fingerprint recognition system for use with low-quality prints and damaged fingertips", Pattern Recognition, vol. 34(2):255–270, 2001.

[44] A.M.Bazen and S.H.Gerez, "Achievement and challenges in fingerprint recognition", in Biometric Solutions for Authentication i an e-World, 2002, pp.23–57.

[45] L. Coetzee and E. C. Botha, "Fingerprint recognition in low quality images", Pattern Recognition, vol. 26(10), 1993, pp. 1441–1460.

[46]   Zhixin Shi , Venu Govindaraju, "A chaincode based scheme for fingerprint feature extraction", Pattern Recognition Letters, vol. 27, 2006, pp. 462–468.

[47]   Zenzo, L. Cinque, and S. Levialdi, "Run-Based Algorithms for Binary Image Analysis and Processing", IEEE Trans. Pattern Analysis and Machine Intelligence, vol. 18, no. 1, 1996, pp. 83-88.

[48]   J Hwan Shin, H. Y. Hwang, S Chien, "Detecting fingerprint minutiae by run length encoding scheme", Pattern Recognition vol. 39, 2005, pp. 1140-1154.

[49]   M. Gamassi, V. Pivri and F. Scotti, "Fingerprint local analysis for high performance minutiae extraction", IEEE International Conference on Image Processing (ICIP) vol. 3, 2005, pp. 265-268.

[50]   E. Alibeigi, M. T. Rizi, P. Behnamfar, "Pipelined minutiae extraction from fingerprint images," Electrical and Computer Engineering, 2009. CCECE '09. Canadian Conference on , vol., no., pp.239-242.

[51]   S. Maddala, S. R. Tangellapally, J. S. Bartuněk and M. Nilsson,   "Implementation and evaluation of NIST Biometric Image Software for fingerprint recognition," Biosignals and Biorobotics Conference (BRC), 2011 ISSNIP, pp.1-5.

[52]   Q. Xiao and H. Raafat, "Fingerprint image postprocessing: a combined statistical and structural approach", Pattern Recognition vol. 24, no. 10, 1991, pp. 985–992.

[53]   J. C. Amengual, A. Juan, J. C. Prez, F. Prat, S. Sez, and J. M. Vilar, "Real-time minutiae extraction in fingerprint images", in Proc. of the 6th Int. Conf. on Image Processing and its Applications, 1997, pp. 871–875.

[54]   A. Farina, Z. M. Kovacs-Vajna, and A. Leone, "Fingerprint minutiae extraction from skeletonized binary images", Pattern Recognition, vol. 32(5), 1999, pp. 877–889.

[55]   J. Xudong and Y. Wei-Yun, "Fingerprint minutiae matching based on the local and global structures", in Proc. of International Conference on Pattern Recognition (ICPR), vol. 2, 2000, pp. 1038–1041.

[56]   M. Tico and P. Kuosmanen, "An algorithm for fingerprint image postprocessing", in Proceedings of the Thirty-Fourth Asilomar Conference on Signals, Systems and Computers, vol. 2, 2000,  pp. 1735–1739.

[57]   S. Prabhakar, A. K. Jain, and S. Pankanti, "Learning fingerprint minutiae location and type", Pattern Recognition, vol. 36(8), 2003, pp. 1847–1857.

[58]   S. Shah, P. S. Sastry, "Fingerprint Classification Using a Feedback Based Line Detector", IEEE Trans. On Systems, Man and Cybernetics, Part B, vol. 34, no.1, 2004.

[59]   S. Chikkerur, V. Govindaraju, S. Pankanti, R. Bolle, and N. Ratha, "Novel approaches for minutiae verification in fingerprint images", in Seventh IEEE Workshops on Application of Computer Vision (WACV/MOTION'05), vol. 1, 2005, pp. 111–116.

[60]   Zhao Feng, Xiaou Tang, "Preprocessing and post processing for skeleton-based   fingerprint minutiae extraction", Pattern Recognition vol. 40, 2007, pp. 1270-1281.

[61]   F. Zhao and X. Tang, "Preprocessing and postprocessing for skeleton-based fingerprint minutiae extraction", Pattern Recognition, vol. 40(4), 2007, pp. 1270–1281.

[62]   M. Usman Akram , A. Tariq, Shoaib A. Khan, " Fingerprint image : pre and post processing", Int. Journal  of Biometrics, Vol. 1, No.1, 2008.

[63]   B. N. Lavanya, K. B. Raja, K. R. Venugopal, L. M. Patnaik, "Minutiae Extraction in Fingerprint Using Gabor Filter Enhancement," In Proc. International Conference on Advances in Computing, Control, & Telecommunication Technologies, ACT '09, 2009, pp.54-56.

[64]   R. Kaur, P. S. Sandhu and A. Kamra, "A Novel Method for Fingerprint Feature Extraction", In Proc. International Conference on Networking and Information Technology, 2010.

[65]   A.R. Patil, M. A. Zaveri, "A Novel Approach for Fingerprint Matching Using Minutiae," In Proc. Fourth Asia International Conference on Mathematical/Analytical Modelling and Computer Simulation (AMS), 2010, pp.317-322.

[66]   P. Pathak, "Image Compression algorithms for Fingerprint System", IJSCI International Journal of Computer Science Issues, vol. 7, issue 3, no. 9, 2010, pp. 45-50.

[67]   P. Gnanasivam and S. Muttan, "An efficient algorithm for fingerprint preprocessing and feature extraction", ICEBT 2010, Procedia computer Science, vol. 2, 2010, pp.133-142.

[68]   W. F. Leung, S. H. Leung, W. H. Lau, A. Luk, "Fingerprint Recognition using Neural Networks", in Proc. IEEE Workshop on Neural  Networks for Signal Processing, 1991, pp. 226-235.

[69]   V. Humbe, S. S. Gornale, R. Manza and K. V. Kale, "Mathematical Morphology approach for Genuine Fingerprint Feature Extraction", Int. Journal of Computer Science and Security (IJCSS), vol. 1, 2007, pp. 53-59.

[70]   R. Bansal, P. Sehgal, P. Bedi, "Effective Morphological Extraction of True Fingerprint Minutiae based on the Hit or Miss Transform", International Journal of Biometrics and Bioinformatics(IJBB), vol. 4, 2010, pp. 71-85.

[71]   D. Maio and D. Maltoni, "Direct gray-scale minutiae detection in fingerprints", IEEE Transactions on Pattern Analysis and Machine Intelligence, 19(1):27–40.

[72]   D. Maio and D. Maltoni, "Neural network based minutiae filtering in fingerprints", in Fourteenth International Conference Pattern Recognition, vol.  2, 1998, pp. 1654–1658.

[73]   X. Jiang, W.-Y. Yau, and W. Ser, "Detecting the fingerprint minutiae by adaptive tracing the gray-level ridge", Pattern Recognition, vol. 34(5), 2001, 999–1013.

[74]   L. Jinxiang, H. Zhongyang, and C. Kap Luk, "Direct minutiae extraction from gray-level fingerprint image by relationship examination", in International Conference on Image Processing(ICIP), vol. 2, 2000, pp. 427–430.

[75]   K. Nilsson and J. Bign, "Using linear symmetry features as a pre-processing step for fingerprint images", in AVBPA, 2001, pp.247–252.

[76]   K. K. Hartwig Fronthaler and J. Bigun, "Local feature extraction in fingerprints by complex filtering. In Advances in Biometric Person Authentication", LNCS, vol. 3781, 2005, pp.77–84.

[77]   X. Gao, X. Chen, J. Cao, C. Deng, C. Liu and J. feng, "A Novel Method Of Fingerprint Minutiae Extraction Based On Gabor Phase", In Proc. IEEE International Conference on Image Processing, 2010, pp. 3077-3080.

[78]  N. K. Ratha, S. Chen, and A. K. Jain, "Adaptive flow orientation-based feature extraction in fingerprint images". Pattern Recognition, vol. 28(11), 1995, pp. 1657–1672.

[79]  V. K. Sagar, D. B. L. Ngo, and K. C. K. Foo, "Fuzzy feature selection for fingerprint identification", in proc. 29th Annual International Carnahan.Security Technology, 1995, pp 85-90.

[80]  V. K. Sagar and K. J. Beng, "Hybrid fuzzy logic and neural network model for fingerprint minutiae extraction", International Joint Conference on Neural Networks, IJCNN '99., vol. 5, 1999, pp. 3255–3259.

[81]  Guang-Ho Cha, "A Context-Aware Similarity Search for a Handwritten Digit Image Database", The Computer Journal, 2009.

[82]  A. K. Jain, Y. Chen, and M. Demirkus, "A fingerprint recognition algorithm combining phase-based image matching and feature-based matching", in Proc. of International Conference on Biometrics (ICB), 2005, pp. 316–325.

[83]  A. K. Jain, Y. Chen, and M. Demirkus, "Pores and ridges: Fingerprint matching using level 3 features", in Proc. of International Conference on Pattern Recognition (ICPR), vol. 4, 2006, pp. 477–480.

[84]  K. Nandakumar and A.K. Jain, "Local Correlation-based Fingerprint Matching", in Proc. ICVGIP, 2004, pp.503-508.

[85]  A. Lindoso, L. Entrena, C. López-Ongil, and J. Liu-Jimenez, "Correlation-Based Fingerprint Matching Using FPGAs", in Proc. FPT, 2005, pp.87-94.

[86]  D. K. Karna, S. Agarwal, and S. Nikam. "Normalized Cross-Correlation Based Fingerprint Matching". In proc. Fifth International Conference on Computer Graphics, Imaging and Visualisation (CGIV '08), 2008, pp. 229-232.

[87]  S. Klein, A. M. Bazen, and R. Veldhuis, "Fingerprint image segmentation based on hidden markov models", in 13th Annual workshop in Circuits, Systems and Signal Processing, 2002.

[88]  F. Alonso-Fernandez, J. Fierrez-Aguilar, and J. Ortega-Garcia, "An enhanced gabor filter-based segmentation algorithm for fingerprint recognition systems", in proc. 4th International Symposium on Image and Signal Processing and Analysis(ISPA 2005), pp. 239–244.

[89]  X. Chen, J. Tian, J. Cheng, and X. Yang, "Segmentation of fingerprint images using linear classifier", EURASIP Journal on Applied Signal Processing, vol. 4, 2004, pp. 480–494.

[90]  E. Zhu, J. Yin, C. Hu, and G. Zhang, "A systematic method for fingerprint ridge orientation estimation and image segmentation", Pattern Recognition, vol. 39(8), 2006, 1452–1472.

[91]  Z. Yuheng and X. Qinghan, "An optimized approach for fingerprint binarization", in International Joint Conference on Neural Networks, 2006, pp. 391–395.

[92]  N. Otsu, "A threshold selection method from gray level histograms", IEEE Transactions on Systems, Man and Cybernetics, vol. 9, No. 1, 1979, pp. 62-66.

[93]  V. Espinos, "Mathematical Morphological approaches for Fingerprint Thinning", in Proc. 36th Annual International Carnahan Conference on Security Technology, 2002, pp. 43-45.

[94]  M. Ahmed and R. Ward, "A rotation invariant rule-based thinning algorithm for character recognition", Pattern Analysis and Machine Intelligence, IEEE Transactions on, vol. 24(12), 2002, pp. 1672–1678.

[95]  P. M. Patil, S. R. Suralkar, and F. B. Sheikh, "Rotation invariant thinning algorithm to detect ridge bifurcations for fingerprint identification", in proc. 17th IEEE International Conference on Tools with Artificial Intelligence, 2005.

[96]  X. You, B. Fang, V. Y. Y. Tang, and J. Huang, " Multiscale approach for thinning ridges of fingerprint", in Proc. Second Iberian Conference on Pattern Recognition and Image Analysis, volume LNCS 3523, 2005, pp. 505–512.

**First Author** Roli Bansal is currently pursuing Ph. D. under the joint supervision of Dr. Punam Bedi and Dr. Priti Sehgal from the Dept. of Computer Science, University of Delhi. She is working in the area of fingerprint image enhancement and watermarking. Earlier, she completed her M.C.A. in 1997 and since then she has been working as Assistant Professor in Keshav College, University of Delhi.

**Second Author Dr. Priti Sehgal** received her Ph.D. in Computer Science from the Department of Computer Science, University of Delhi, India in 2006 and her M. Sc. in Computer Science from DAVV, Indore, India in 1994. She is an Associate Professor in the Department of Computer Science, Keshav Mahavidyalaya, University of Delhi. She has about 17 years of teaching and research experience and has published papers in National/International Journals/Conferences. Dr. Sehgal has been a member of the program committee of the CGIV International Conference and is a life member of Computer Society of India. Her research interests include Computer Graphics, Image Processing, Biometrics, Visualization and Image Retrieval.

**Third Author Dr. Punam Bedi** received her Ph.D. in Computer Science from the Department of Computer Science, University of Delhi, India in 1999 and her M.Tech. in Computer Science from IIT Delhi, India in 1986. She is an Associate Professor in the Department of Computer Science, University of Delhi. She has about 25 years of teaching and research experience and has published about 110 papers in National/International Journals/Conferences. Dr. Bedi is a member of AAAI, ACM, senior member of IEEE, and life member of Computer Society of India.
Her research interests include Web Intelligence, Soft Computing, Semantic Web, Multi-agent Systems, Intelligent Information Systems, Intelligent Software Engineering, Intelligent User Interfaces, Requirement Engineering, Human Computer Interaction (HCI), Trust, Information Retrieval and Personalization.